%% file: MAIN.tex
\documentclass[10pt,conference]{IEEEtran}
\IEEEoverridecommandlockouts

\usepackage{cite}
\usepackage{amsmath,amssymb,amsfonts}
\usepackage{algorithmic}
\usepackage{graphicx}
\usepackage{textcomp}
\usepackage{xcolor}
\usepackage{multirow}
\usepackage{booktabs}
\def\BibTeX{{\rm B\kern-.05em{\sc i\kern-.025em b}\kern-.08em
    T\kern-.1667em\lower.7ex\hbox{E}\kern-.125emX}}
    
\begin{document}

\title{SUGAR: Leveraging Contextual Confidence for Smarter Retrieval}
\author{\IEEEauthorblockN{1\textsuperscript{st} Hanna Zubkova} \
\IEEEauthorblockA{\textit{Department of Artificial Intelligence} \\
\textit{Korea University}\\
Seoul, Korea \\
zubkova{\_}hanna@korea.ac.kr}
\and
\IEEEauthorblockN{2\textsuperscript{nd} Ji-Hoon Park}
\IEEEauthorblockA{\textit{Department of Artificial Intelligence} \\
\textit{Korea University}\\
Seoul, Korea \\
jhoon{\_}park@korea.ac.kr}
\and
\IEEEauthorblockN{3\textsuperscript{rd} Seong-Whan Lee$^\dagger$ \thanks{$^\dagger$Corresponding author.}}
\IEEEauthorblockA{\textit{Department of Artificial Intelligence} \\
\textit{Korea University}\\
Seoul, Korea \\
sw.lee@korea.ac.kr}
}

\maketitle

\begin{abstract}
Bearing in mind the limited parametric knowledge of Large Language Models (LLMs), retrieval-augmented generation (RAG) which supplies them with the relevant external knowledge has served as an approach to mitigate the issue of hallucinations to a certain extent. However, uniformly retrieving supporting context makes response generation source-inefficient, as triggering the retriever is not always necessary, or even inaccurate, when a model gets distracted by noisy retrieved content and produces an unhelpful answer. Motivated by these issues, we introduce \textbf{Semantic Uncertainty Guided Adaptive Retrieval (SUGAR)}, where we leverage context-based entropy to actively decide whether to retrieve and to further determine between single-step and multi-step retrieval. Our empirical results show that selective retrieval guided by semantic uncertainty estimation improves the performance across diverse question answering tasks, as well as achieves a more efficient inference. 
\end{abstract}

\begin{IEEEkeywords}
large language models, retrieval augmented generation, uncertainty estimation, question answering.
\end{IEEEkeywords}

\section{Introduction}
\label{sec:intro}
Despite showing impressive performance results \cite{C1, C3, C33, C34}, recent state-of-the-art Large Language Models (LLMs) still face challenges in tackling knowledge-intensive tasks like open-domain question answering (QA) \cite{C4}. Their generations solely depend on parametric memory of the models, and LLMs lack domain-specific and up-to-date world knowledge, which leads to factual errors in solving QA tasks. Recently, retrieval-augmented generation (RAG) has become a widely applied approach to address this issue, as it provides LLMs with relevant supporting context from an external source \cite{C5, C31, C35, C36}.

Even though RAG clearly helps with mitigating hallucinations, it has some challenges of its own. Namely, it is obviously not necessary to conduct retrieval for every QA case at hand. RAG does help LLMs generate factually accurate outputs when they lack relevant knowledge, but a lot of simpler queries can be answered with just the parametric knowledge of the model, so naively retrieving for every iteration makes inference inefficient \cite{C6}. Moreover, the retrieved results sometimes contain documents that are irrelevant\cite{C7}, factually incorrect \cite{C21} or even contain harmful information \cite{C9}. Recent studies have investigated the knowledge preference between parametric knowledge and external context presented in RAG \cite{C10}. Some works \cite{C11, C12, C15, C39, C40, C41, C42, C43} have shown that LLMs get easily distracted by noisy retrieved documents and generate seemingly plausible, but incorrect outputs, even though the parametric knowledge of the model would have been enough to accurately answer the question, had retrieval not been triggered.

The problem of achieving a harmonious synthesis of external and internal knowledge within LLMs has inspired a whole line of RAG research that focuses on the question \textit{``when to retrieve?"}. Adaptive RAG \cite{C6} dynamically decides whether to retrieve based on class labels that reflect question complexity, Self-RAG \cite{C13} uses self-reflection tokens which signal the need for retrieval or confirm the output relevance, support, or completeness. UniWeb \cite{C17} retrieves only in the case of small predictive entropy of the output distribution. FLARE \cite{C16} retrieves relevant documents if a prediction of the upcoming sentence contains any low-confidence tokens. For complex multi-hop questions, IRCoT \cite{C23} has been proposed to iteratively interact with the both LLM and the retriever. However, such uniform multi-step retrieval becomes very resource-intensive or heavily data-dependent, and calculating naive entropy has a major downside unique to the area of natural language processing — in language generation the same output can be produced in a variety of linguistic forms \cite{C37}. So, when calculating predictive entropy without accounting for meaning, different surface forms compete for probability mass, even if they represent the same idea \cite{C18}. As a result, models either get confused by variations of the same meaning or, vice versa, exhibit overconfidence when lacking relevant knowledge and generating semantically disperse answers.

\begin{figure*}[h]

\centering
    \includegraphics[width=\textwidth]{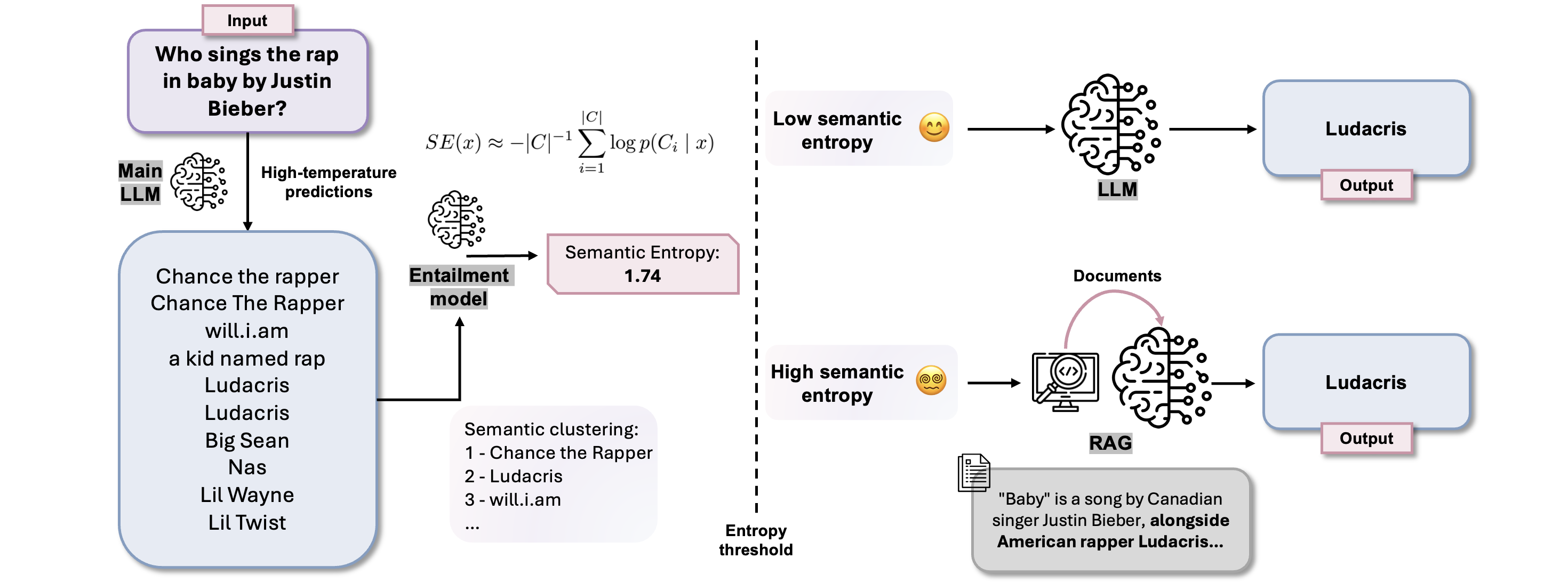}
    \caption{Overview of the proposed retrieval strategy. Semantic entropy is used to measure how confident the model is to answer the question based on its parametric knowledge. (A) If semantic entropy is low, the answer is generated based on internal knowledge, (B) if semantic entropy is high, the retriever is triggered to find relevant external knowledge, which is used to generate the answer.}
    \label{fig:main_figure}

\end{figure*}

To address these points, in contrast with previous works and inspired by Kuhn et al. \cite{C19}, we propose using semantic entropy as the defining metric for whether to conduct retrieval or not. With \textbf{Semantic Uncertainty Guided Adaptive Retrieval (SUGAR)}, our intuition is that accounting for linguistic invariances improves the knowledge boundary evaluation of LLMs in general. Therefore, we believe that if provided with external context when models are uncertain of generating their answers, semantic uncertainty would make retrieval more controllable. This would improve the overall quality of QA performance by triggering the retriever only when it is necessary. 

In summary, our contributions are as follows: (1) We propose SUGAR, an adaptive RAG strategy based on semantic uncertainty, which dynamically decides whether to conduct single-step retrieval, multi-step retrieval or to not conduct retrieval at all; our approach does not require additional training or fine-tuning, and is not task- or data- dependent. (2) We validate the proposed retrieval strategy using benchmark single- and multi-hop open-domain QA datasets, and empirically show that semantic entropy in SUGAR is effective to determine whether retrieval is necessary, supports robust performance, and helps mitigate overconfidence.

\section{SUGAR: Semantic Uncertainty Guided Adaptive Retrieval}
\label{sec:method}

In this section, we first outline the overall Semantic Uncertainty Guided Adaptive Retrieval (SUGAR) framework in Section 2.1, and then introduce the proposed strategy in detail. Specifically, in Section 2.2, we explain the idea behind Semantic Uncertainty; we then advocate for its adaptation in adaptive retrieval-augmented generation in Section 2.3. 

\subsection{Framework overview}
\label{ssec:overview}
As presented in Figure~\ref{fig:main_figure}, given a question \textit{q}, we use semantic entropy to evaluate how uncertain the model is of generating an answer \textit{a} with respect to \textit{q} using just its parametric knowledge \textit{P}. We set a confidence threshold $\tau$, and if the model is confident in its output, it simply proceeds with generating the answer \textit{a}. However, in the case of exhibiting high semantic uncertainty, we call the retriever. External knowledge \textit{D} is extracted and used as supporting context to generate the answer \textit{a}. When conducting retrieval, we propose using the confidence threshold $\tau$ to dynamically decide between single-step and multiple-step retrieval. 

\subsection{Semantic Uncertainty}
\label{ssec:semantic_uncertainty}
One of the challenges of using entropy for uncertainty estimation in free-form language generation is that, unlike in other machine learning problems, where outputs are mutually exclusive, in natural language generation we can express the same idea in a variety of syntactic and lexical forms. Regular predictive entropy does not account for this fact, as it is computed based on token-likelihoods. To address this, Kuhn et al. \cite{C19} instead propose using \textit{semantic} entropy,

\begin{equation}
    SE(x) \approx -|C|^{-1} \sum_{i=1}^{|C|} \log p(C_i \mid x)
    \label{eq:1}
\end{equation}

\input{table1}
\input{table3}

to compute model uncertainty after clustering together sequences that vary in lexical form but still carry the same meaning, i.e. semantic value. First, 10 potential high-temperature answers are generated, then bidirectional entailment is used to detect varying forms of one meaning among these generations – namely, two sequences mean the same thing if they logically imply each other. Lastly, the likelihood of generating each of the semantic clusters \textit{C} is computed (in contrast to regular predictive entropy, which would consider the likelihood of each individual sequence separately). The experimental results demonstrate, that entropy, which accounts for semantic value, indeed measures uncertainty better than the regular predictive entropy. Inspired by such conclusion, we follow the proposed approach and argue for its adaptation as a more fit LLM uncertainty estimation metric in the context of selective retrieval.

\subsection{Adaptive Retrieval Augmented Generation}
\label{ssec:adaptive_RAG}
Adequate evaluation of knowledge boundaries in language models is crucial to make retrieval more controllable and efficient. In LLM uncertainty estimation, for black-box models, it has been common to prompt the model itself to judge its own ability \cite{C14}, but it is infeasible to estimate how truthful and faithful LLMs are when answering questions like ``Can you answer this question? It this answer correct?" On the other hand, logit-based uncertainty estimation methods, even though not applicable to black-box models, essentially seem more reliable as they can be quantified and measured. Such methods, like predictive entropy or semantic entropy, are being actively used in the line of research that focuses on detecting hallucinations in LLMs and eliciting abstention \cite{C38}. 

It is important, though, that a model might be highly uncertain between generating something like ``Shakespeare wrote Romeo and Juliet" or ``Romeo and Juliet was written by Shakespeare" token-by-token and therefore exhibit high entropy, as these sequences differ in terms of form, but convey the same idea. Since in knowledge intensive tasks like QA we care about factual accuracy, the lexical form does not really matter as long as the answer is correct. Semantic entropy supports the generation process when the model is confused by subtle lexical variations, and is a clearer indicator of uncertainty when potential answers drastically vary in meaning.

To the best of our knowledge, semantic uncertainty estimation methods have not been implemented in the context of information retrieval, so we advocate for its application as a metric for selective retrieval. With SUGAR we aim to dynamically decide when and how often to retrieve based on semantic entropy thresholds. As computing semantic entropy can be applied to any QA dataset, our approach also suggests a broader generalization potential, compared to the previous methods, namely highly task-dependent reflection tokens in Self-RAG (authors intend to simply retrieve more often for all knowledge intensive tasks), and highly data-dependent complexity labels in Adaptive-RAG (there is no annotated data available to properly train a complexity classifier).

Therefore, we propose to set semantic entropy thresholds that make three uncertainty level intervals with three corresponding retrieval scenarios – `no retrieval' for the lowest semantic entropy (the model is most certain, retrieval is likely unnecessary), `single-step retrieval' for intermediate levels of semantic entropy (the model is somewhat uncertain, one round of retrieval would help with answer generation), and `multi-step retrieval' for the highest entropy (the model is highly uncertain, multiple rounds of retrieval would be helpful).

\section{Experiments and Results}
\label{sec:experiments}

\subsection{Datasets and metrics}
\label{ssec:datasets}
We evaluate the proposed strategy on the classic single-hop open-domain QA datasets: SQuAD \cite{C27}, Natural Questions \cite{C28} and TriviaQA \cite{C20}. To evaluate how the method performs on more complex questions, we also use the following multi-hop datasets: HotpotQA \cite{C25} and 2WikiMultiHopQA \cite{C26}. We use accuracy, F1, and EM to measure effectiveness, and the number of retrieval steps and answering time relative to single-step retrieval to measure efficiency.

\subsection{Baselines}
\label{ssec:baselines}
We use the off-the-shelf version of FLAN-T5-XL \cite{C30} for the \textbf{No Retrieval} setting, and the same model augmented with a retriever for \textbf{Single-step} retrieval. We also compare SUGAR to previously proposed \textbf{Adaptive} retrieval strategies: Adaptive Retrieval \cite{C4} based on entity popularity, Self-RAG \cite{C13} based on reflection tokens, and Adaptive-RAG \cite{C6} based on question complexity labels. Additionally, we consider IRCoT \cite{C23}, which accesses the retriever and the generator with interleaving Chain-of-Thought reasoning, as the \textbf{Multi-hop Retrieval} baseline. As analyzed in the Adaptive-RAG paper, their method performs well on multi-hop datasets primarily due to the direct integration of the IRCoT strategy, which neither our approach nor other baselines utilize. Thus, for fair comparison on the multi-hop datasets, we exclude Adaptive-RAG. We run our experiments in a one-shot manner with one task demonstration of the ``Q: $<$question$>$ A:" format.

\begin{figure}[h]

\centering
    \includegraphics[width=\columnwidth]{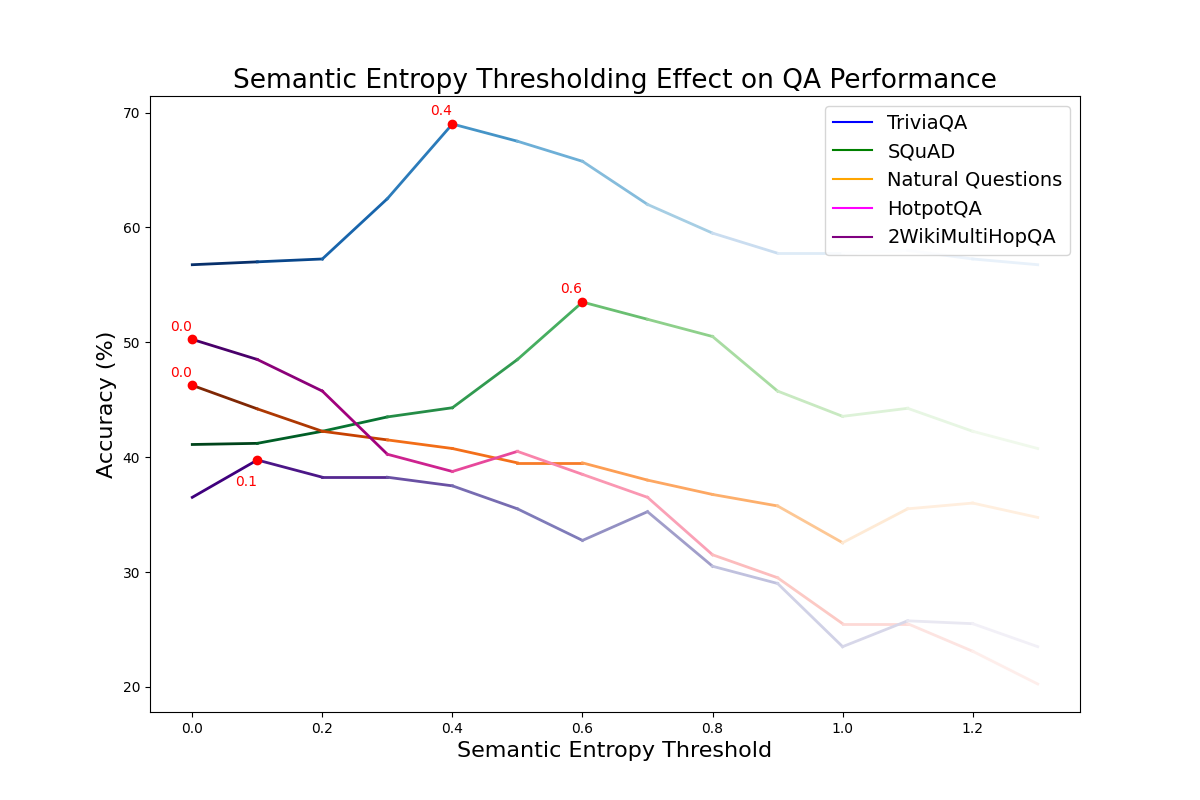}
    \caption{Semantic entropy levels and corresponding accuracy. Gradient indicates retrieval frequency (as the color fades out, retrieval is triggered less), we mark the semantic entropy levels used as thresholds in red.}
    \label{fig:thresholds}

\end{figure}

\subsection{Results}
\label{ssec:results}
In SUGAR, we use Llama-2-chat (7B) \cite{C29} as the generator and off-the-shelf Contriever-MS MARCO \cite{C31} as the retriever. To set the semantic entropy thresholds, we first performed a case study on the datasets to see what levels of semantic entropy the model normally demonstrates when generating answers. We then used cross-validation to determine the thresholds that yield the best performance in terms of effectiveness metrics, we report the case study results in Figure~\ref{fig:thresholds}.

Tables~\ref{tab:table1} and~\ref{tab:table3} summarize our primary results, showing improvements for both effectiveness and efficiency, compared to naive single-step and multi-step retrieval. With SUGAR outperforming the other adaptive approaches effectiveness-wise, we can also see the positive effect of using semantic entropy as the retrieval trigger. For the multi-hop datasets, SUGAR outperforms its baselines in terms of the number of retrieval steps, while still showing superior accuracy performance. 

Similarly to Adaptive-RAG, semantic entropy intervals allow for a fine-grained treatment of various questions, but in our approach we can estimate how `confused' a certain input makes the generator model without depending on training a classifier model. In this regard, we believe that a drawback of our method is time-dependency. While efficiently reducing the number of necessary retrieval steps for both single-hop and multi-hop datasets, we faced a trade-off in terms of inference time. As it is necessary to compute semantic entropy, inference for SUGAR takes longer than other adaptive methods for single-hop datasets. Notably, however, SUGAR is still consistently faster than IRCoT even when multiple retrieval rounds are done, and the inference time also does not significantly increase when switching to the multi-hop datasets.

\input{table2}

\subsection{Ablation Study and Analyses}
\label{ssec:results}

To further validate the effect of semantic entropy, we additionally compared Llama-2-chat (7B) without retrieval, uniform single-step retrieval, adaptive retrieval based on regular predictive entropy, and adaptive retrieval based on semantic entropy (SUGAR). As mentioned before, datasets naturally vary in complexity, so we observed the average values of predictive and semantic entropy for each dataset we experimented on. Representatively, for TriviaQA, predictive entropy was consistently higher than semantic entropy, while the opposite was the case for SQuAD. To demonstrate both possible scenarios of entropy variations, for the following analysis we decided to compare the performance on these two datasets. 

In this experiment we set the single-hop SUGAR thresholds to the average values for both datasets ($\tau$ = 0.4 for TriviaQA, $\tau$ = 0.9 for SQuAD) And for fair comparison, for predictive entropy we set two possible thresholds – average predictive entropy values ($\tau$ = 0.55 for TriviaQA, $\tau$ = 0.7 for SQuAD), and thresholds equal to the ones used for SUGAR ($\tau$ = 0.4 for TriviaQA, $\tau$ = 0.9 for SQuAD). We compare answer accuracy for effectiveness, and for efficiency the number of retrieval steps is averaged over both datasets, we report the ablation study results in Table~\ref{tab:table2}.

Semantic entropy being consistently higher than regular predictive entropy points out the variability in potential answers and suggests the model is being overconfident (lower predictive entropy combined with high lexical variance in potential answers leads to believe model predictions might not be as reliable). In the opposite case when semantic entropy is lower, while the model is less certain about specific outputs, these outputs are semantically consistent and convey the same idea. But notably, for both datasets, we can see that context-sensitive semantic entropy performs much better and stays more efficient than regular predictive entropy, which leads us to believe that not only does semantic entropy help mitigate overconfidence, it also supports robust performance when the model encounters slight lexical form variations.

\section{Conclusion}
\label{sec:conclusion}

In this work we proposed Semantic Uncertainty Guided Adaptive Retrieval, which we refer to as SUGAR, to dynamically determine the necessity of retrieving external knowledge in open-domain QA. The main idea of SUGAR is to use semantic entropy to assess if the parametric knowledge is sufficient to answer a question, and retrieve supporting external context if it is not. Semantic entropy is fit for free-form language generation as it estimates LLMs uncertainty over meaning to evaluate its knowledge boundaries. Our results show that SUGAR improves overall accuracy on QA tasks and proves to be an efficient retrieval strategy which allows the combination of using parametric and external knowledge.

\section*{Acknowledgments}
\label{sec:acknowledgments}
This work was partly supported by the Institute of Information \& Communications Technology Planning \& Evaluation (IITP) grant funded by the Korea government (MSIT) (No. RS-2019-II190079, Artificial Intelligence Graduate School Program (Korea University), No. RS-2024-00336673, AI Technology for Interactive Communication of Language Impaired Individuals, and No. RS-2024-00436857, Information Technology Research Center (ITRC) support program).

\bibliographystyle{IEEEtran}
\bibliography{refs}

\end{document}

%% file: table1.tex
\begin{table*}[h!]
\caption{Experiment results on Single-hop QA datasets.}
\label{tab:table1}
\normalsize
\resizebox{\textwidth}{!}{
\begin{tabular}{lllllllllllllllllllllll}
\toprule
\multirow{2}{*}{Data} & \multirow{2}{*}{Types} & \multirow{2}{*}{Methods} & \multicolumn{5}{c}{SQuAD} & \multicolumn{5}{c}{Natural Questions} & \multicolumn{5}{c}{TriviaQA} \\ \cmidrule(r){4-18}
& & & \multicolumn{1}{c}{EM} & \multicolumn{1}{c}{F1} & \multicolumn{1}{c}{Acc} & \multicolumn{1}{c}{Step} & \multicolumn{1}{c}{Time} & \multicolumn{1}{c}{EM} & \multicolumn{1}{c}{F1} & \multicolumn{1}{c}{Acc} & \multicolumn{1}{c}{Step} & \multicolumn{1}{c}{Time} & \multicolumn{1}{c}{EM} & \multicolumn{1}{c}{F1} & \multicolumn{1}{c}{Acc} & \multicolumn{1}{c}{Step} & \multicolumn{1}{c}{Time} \\ \midrule

\multirow{8}{*}{Single-hop} 
& \multirow{2}{*}{Simple} & No Retrieval & 3.60 & 10.50 & 5.00 & 0.00 & 0.11 & 14.20 & 19.00 & 15.60 & 0.00 & 0.13 & 25.00 & 31.80 & 27.00 & 0.00 & 0.13 \\
& & Single-step Approach & 27.80 & 39.30 & 34.00 & 1.00 & 1.00 & 37.80 & 47.30 & 44.60 & 1.00 & 1.00 & 53.60 & 62.40 & 60.20 & 1.00 & 1.00 \\ \cmidrule(r){2-18}

& \multirow{4}{*}{Adaptive} & Adaptive Retrieval \cite{C4} & 13.40 & 23.10 & 17.60 & 0.50 & 0.55 & 28.20 & 36.00 & 33.00 & 0.50 & 0.56 & 38.40 & 46.90 & 42.60 & 0.50 & 0.56 \\
& & Self-RAG \cite{C13} & 2.20 & 11.20 & 18.40 & 0.63 & 0.50 & 31.40 & 39.00 & 33.60 & 0.63 & 0.17 & 12.80 & 29.30 & 57.00 & 0.68 & 0.45 \\
& & Adaptive-RAG \cite{C6} & 26.80 & 38.30 & 33.00 & 1.37 & 2.02 & 37.80 & 47.30 & 44.60 & 1.00 & 1.00 & 52.20 & 60.70 & 58.20 & 1.23 & 1.54 \\
& & SUGAR (Ours) & \textbf{34.50} & \textbf{47.19} & \textbf{53.50} & \textbf{1.23} & 4.43 & 31.75 & 41.30 & \textbf{46.25} & \textbf{1.00} & \textbf{1.00}  & \textbf{55.75} & \textbf{64.25} & \textbf{69.25} & \textbf{0.77} & 3.13 \\ \cmidrule(r){2-18}

& \multirow{1}{*}{Complex} & Multi-step Approach \cite{C23} & 24.40 & 35.60 & 29.60 & 4.52 & 9.03 & 38.60 & 47.80 & 44.20 & 5.04 & 10.18 & 53.80 & 62.40 & 60.20 & 5.28 & 9.22 \\ \midrule

\end{tabular}
}
\end{table*}

%% file: table3.tex
\begin{table}[t]
\caption{Experiment results on Multi-hop QA datasets.}
\label{tab:table3}
\centering
\resizebox{\columnwidth}{!}{  
\begin{tabular}{ccccccccccc}
\toprule
\multirow{2}{*}{Methods} & \multicolumn{5}{c}{HotpotQA} & \multicolumn{5}{c}{2WikiMultiHopQA} \\ \cmidrule(lr){2-6} \cmidrule(lr){7-11}
& EM & F1 & Acc & Step & Time & EM & F1 & Acc & Step & Time \\ \midrule
No Retrieval & 16.60 & 22.71 & 17.20 & 0.00 & 0.11 & 27.40 & 32.04 & 27.80 & 0.00 & 0.10 \\
Single-step & 34.40 & 46.15 & 36.40 & 1.00 & 1.00 & 41.60 & 47.90 & 42.80 & 1.00 & 1.00 \\ \midrule

Adaptive Retrieval \cite{C4} & 23.60 & 32.22 & 25.00 & 0.50 & 0.55 & 33.20 & 39.44 & 34.20 & 0.50 & 0.55 \\
Self-RAG \cite{C13} & 6.80 & 17.53 & 29.60 & 0.73 & 0.45 & 4.60 & 19.59 & 38.80 & 0.93 & 0.49 \\
SUGAR (Ours) & \textbf{38.77} & \textbf{49.85} & \textbf{50.25} & \textbf{2.31} & \textbf{5.11} & \textbf{25.00} & \textbf{41.92} & \textbf{39.75} & \textbf{0.88} & \textbf{4.86} \\ \midrule

Multi-step \cite{C23} & 44.60 & 56.54 & 47.00 & 5.53 & 9.38 & 49.60 & 58.85 & 55.40 & 4.17 & 7.37 \\ \bottomrule
\end{tabular}
}
\end{table}

%% file: table2.tex
\begin{table}[t]
\caption{Ablation study results.}
\label{tab:table2}
\normalsize
\resizebox{\columnwidth}{!}{
\begin{tabular}{c c c c}
\toprule
Methods & TriviaQA Acc & SQuAD Acc & Step \\ \midrule
No retrieval & 55.50 & 18.25 & 0.00 \\
Single-step retrieval & 58.50 & 22.75 & 1.00 \\
Predictive entropy (average $\tau$) & 65.00 & 34.25 & 1.04 \\
Predictive entropy (semantic $\tau$) & 67.75 & 36.75 & 1.36 \\ \midrule
SUGAR (ours) & \textbf{69.00} & \textbf{50.75} & \textbf{0.91} \\ \bottomrule
\end{tabular}
}
\end{table}